\documentclass{article}
\usepackage{spconf,amssymb,amsmath,graphicx,epstopdf,comment,subfig,algorithm}
\usepackage{bm}
\usepackage{comment}
\usepackage{balance}
\usepackage{units}
\usepackage[nospace]{cite}
\usepackage{relsize}
\usepackage{multicol}

\graphicspath{{Figures/}}

\def \T{\mathsf{T}}

\def \0{\mathbf{0}}

\newcommand*{\E}[1]{\mathsf{E} \left\{ #1 \right\}}

\def \T{\mathsf{T}}

\newcommand*{\diag}[1]{\mathsf{diag}\left\{#1\right\}}

\newcommand*{\p}[1]{\mathsf{p} \left\{ #1 \right\}}

\newcommand*{\tr}[1]{\mathsf{tr} \left\{ #1 \right\}}
\usepackage{relsize}

\DeclareMathAlphabet\mathbfcal{OMS}{cmsy}{b}{n}

\title{On the Dynamics of Multiagent Nonlinear filtering and Learning}
\name{Sayed Pouria Talebi$^\dagger$ and Danilo P. Mandic$^\ddagger$}
\address{$^\dagger$Computer Science Department, University of Roehampton, London, SW15~5PH, UK \\$^\ddagger$Electrical and Electronic Engineering Department, Imperial College London, SW7~2AZ, UK
	\\E-mails: sayed.talebi@roehampton.ac.uk, \& d.mandic@imperial.ac.uk}
\begin{document}
\ninept
\maketitle
\begin{abstract}
Multiagent systems aim to accomplish highly complex learning tasks through decentralised consensus seeking dynamics and their use has garnered a great deal of attention in the signal processing and computational intelligence societies. This article examines the behaviour of multiagent networked systems with nonlinear filtering/learning dynamics. To this end, a general formulation for the actions of an agent in multiagent networked systems is presented and conditions for achieving a cohesive learning behaviour is given. Importantly, application of the so derived framework in distributed and federated learning scenarios are presented. 
\end{abstract}

\begin{keywords}
Multiagent systems, nonlinear dynamics, distributed learning, federated learning, 
\end{keywords}

\section{Introduction}


Traditionally, signal processing and  learning techniques have been concerned with single agent operations~\cite{LinearEstimation,DB}. However, as early as the $1970$s, it became clear that these agents will come into contact with one another and in most cases cooperation will be beneficial~\cite{FirstLQGR,FirstSurvay}. Moreover, nature provides exquisite examples of achieving highly complex behaviour from a set of simple, but collaborating agents~\cite{AliNature,Flocking}. In broad terms, at the heart of most signal processing and learning techniques lie iterative optimisation approaches that aim to minimise a performance metric or cost function. Traditionally, these solutions were formulated with a single information processing unit in mind, a framework henceforth referred to as centralised, as the optimisation procedures are performed at a central information processing unit. Although shown to be effective, these techniques cannot accommodate modern applications that require interaction of multiple agents over large-scale networks~\cite{SayedBook,MARL-Survay}.  

Initial distributed signal processing and learning techniques were based on  the framework of Pareto optimisation~\cite{SayedBook,AliDistLearningOne} and consensus seeking procedures~\cite{OriginalConsensus,NonlinearConsensus}. In the learning arena, the federated learning approach, initiated at Google~\cite{FedLearnOriginal}, has gained traction. In this approach, agents update the parameter set learnt using information available to them locally; then, they pass these locally updated models to a fusion centre which integrates all the updates and passes back to the agents a global update~\cite{FedLearnOriginal,FedLearn}. In turn, federated learning approaches are built upon distributed optimisation techniques derived for multiagent signal processing tasks~\cite{DistL,FogLearning,AliDistLearningOne}, where agents establish a cohesive signal processing procedure among themselves, using local communications only, without requiring a fusion centre.

The authors have addressed behaviour of linear distributed filtering techniques in~\cite{DistRiccati,OnDistFilter}, proving that convergence conditions of distributed and centralised filtering techniques are similar. This article considers the problem of federated/distributed learning with nonlinear dynamics. To this end, a general class of adaptive learning techniques is formulated. Then, operations of these formulated learning technique is decomposed to derive its federated/distributed dual. Considering discrepancy between performance of centralised and federated/distributed formulations as a metric, convergence criteria of the federated/distributed adaptive learning techniques with nonlinear dynamics is examined. The results significantly relax convergence criteria for distributed iterative optimisation techniques; thus, opening the opportunity for using distributed adaptive learning techniques in new applications areas.  

\vspace{0.08cm}

\noindent\textbf{\textit{Nomenclature}}: Scalars, column vectors, and matrices are denoted by lowercase, bold lowercase, and bold uppercase letters, while remainder of the nomenclature is summarised as follows:

\vspace{0.08cm}

\begin{tabular}{ll}
$\mathbb{N}$ & set of natural numbers
\\
$\mathcal{N}$ & set of all agents in the network
\\
$\mathcal{N}_{i}$ & set of agents that can communicate with agent $i$
\\ 
$|\cdot|$ & cardinality operator
\\
$\|\cdot\|$ & Euclidean norm
\\
$\E{\cdot}$ & statistical expectation operator
\\
 $(\cdot)^{\T}$ & transpose operator
\\
$\delta(\cdot)$ & Kronecker delta function
\\
$\mathbf{I}$ & identity matrix of appropriate size
\\
$\mathbf{1}$ & vector of appropriate size with unit entries
\\
$\otimes$ & Kronecker product
\\
$\diag{\cdot}$ & constructs a block-diagonal matrix from its entries
\\
$\p{\cdot}$ & spectral radius operator
\\
$\tr{\cdot}$ & trace operator
\\
$\nabla_{\chi}$ & gradient operator with respect to $\chi$ 
\end{tabular}

\vspace{0.08cm}


\subsection{Problem Formulation}

The objective is to estimate/track a vector sequence $\{\mathbf{x}_{n}:n\in\mathbb{N}\}$ representing the condition of a system\footnote{This system can range from a physical system to the state of learning machines, such as weights of a neural network.}, that evolves as 
\begin{equation}
\mathbf{x}_{n+1}=\mathrm{f}(\mathbf{x}_{n},\mathbf{v}_{n})
\label{eq:State-Model}
\end{equation}
where $\mathrm{f}(\cdot)$ is an evolution function with $\{\mathbf{v}_{n}:n\in\mathbb{N}\}$ representing  the evolution noise. To this end, a set of agents are used that obtain observations modelled as
\begin{equation}
\forall i\in\mathcal{N}:\hspace{0.16cm}\mathbf{y}_{i,n}=\mathrm{h}_{i}(\mathbf{x}_{n})+\mathbf{w}_{i,n}
\label{eq:Obsrvation-Model}
\end{equation}  
where $\mathbf{y}_{i,n}$ and $\mathbf{w}_{i,n}$ are the observation and observation noise at agent $i$ at time instant $n$, whereas $\mathrm{h}_{i}(\cdot)$ represents the observation function of agent $i$. 

\vspace{0.08cm}

\noindent\textbf{Assumption~1:} Observation and state evolution noises are zero-mean white Gaussian random sequences with joint covariance matrix given by
\begin{equation}
\hspace{-0.008cm}\E{\begin{bmatrix}\mathbf{v}_{n}\\\mathbf{w}_{l,n}\end{bmatrix}\begin{bmatrix}\mathbf{v}_{m}\\\mathbf{w}_{k,m}\end{bmatrix}^{\T}}=\begin{bmatrix}\boldsymbol{\Sigma}_{\mathbf{v}}&\mathbf{0}\\\mathbf{0}&\boldsymbol{\Sigma}_{\mathbf{w}_{l}}\delta(l-k)\end{bmatrix}\delta(n-m).
\label{eq:cov-info}
\end{equation}

\vspace{0.08cm}

On the most fundamental level, this problem can be addressed by organising all observations into a vector, so that 
\[
\underbrace{\begin{bmatrix}\mathbf{y}_{1,n}\\ \vdots\\ \mathbf{y}_{|\mathcal{N}|,n}\end{bmatrix}}_{\mathbf{y}_{n}}=\underbrace{\begin{bmatrix}\mathrm{h}_{1}(\mathbf{x}_{n})\\ \vdots\\ \mathrm{h}_{|\mathcal{N}|}(\mathbf{x}_{n})\end{bmatrix}}_{\mathrm{h}(\mathbf{x}_{n})}+\underbrace{\begin{bmatrix}\mathbf{w}_{1,n}\\ \vdots \\ \mathbf{w}_{|\mathcal{N}|,n}\end{bmatrix}}_{\mathbf{w}_{n}}
\]
which allows for the use of centralised adaptation operations~\cite{Simo}
\begin{equation}
\hat{\mathbf{x}}_{n}=\mathrm{f}(\hat{\mathbf{x}}_{n-1})+\mathbf{G}_{n}\Big(\mathbf{y}_{n}-\mathrm{h}\left(\mathrm{f}\left(\hat{\mathbf{x}}_{n-1}\right)\right)\Big)
\label{eq:Central-Filter}
\end{equation} 
where $\mathbf{G}_{n}$ is a gain matrix calculated to minimise an error measure using a gradient-based format and $\hat{\mathbf{x}}_{n}$ denotes the estimate of $\mathbf{x}_{n}$, that is, $\hat{\mathbf{x}}_{n}=\E{\mathbf{x}_{n}|\mathbf{y}_{1},\ldots,\mathbf{y}_{n}}$, representing the best approximation of the expected value of $\mathbf{x}_{n}$ given the available information, while $\mathrm{f}(\hat{\mathbf{x}}_{n-1})$ is projection of the estimate form time instant $n-1$ onto time instant $n$.

\vspace{0.08cm}

\noindent\textbf{Assumption~2:}
The estimation error of the centralised filter in \eqref{eq:Central-Filter} at time instant $n$ is defined as
\begin{equation}
\boldsymbol{\epsilon}_{n}=\mathbf{x}_{n}-\hat{\mathbf{x}}_{n}
\label{eq:Cent-Error}
\end{equation}
where it is assumed the matrix sequence $\{\mathbf{G}_{n}:n\in\mathbb{N}\}$ exist so that the filtering operations in \eqref{eq:Central-Filter} become exponentially bounded in the mean square error (MSE) sense. That is, for some $\mu>0$ there is $ \eta,\nu>0$ and $\rho\in(0,1)$ so that 
\begin{equation}
\text{if}\hspace{0.12cm}\|\boldsymbol{\epsilon}_{1}\|^{2}\leq\mu\text{; then,}\hspace{0.12cm} \E{\|\boldsymbol{\epsilon}_{n}\|^{2}}\leq\eta\|\boldsymbol{\epsilon}_{1}\|^{2}\rho^{n}+\nu.
\label{eq:Exponetially-Bounded}
\end{equation}

\subsection{From Centralised to Federated Learning}

Focusing on the expression in \eqref{eq:Central-Filter} and partitioning the gain matrix into appropriate sections so that one has 
\[
\mathbf{G}_{n}=\begin{bmatrix}\mathbf{G}_{1,n},\ldots,\mathbf{G}_{|\mathcal{N}|,n}\end{bmatrix}
\]
allows \eqref{eq:Central-Filter} to be expressed as 
\begin{align}
\hat{\mathbf{x}}_{n}=&\mathrm{f}(\hat{\mathrm{x}}_{n-1})+\sum_{\forall i\in\mathcal{N}}\mathbf{G}_{i,n}\Big(\mathbf{y}_{i,n}-\mathrm{h}_{i}\left(\mathrm{f}\left(\hat{\mathbf{x}}_{n-1}\right)\right)\Big)\label{eq:Midway-Term}
\\
=&\underbrace{\frac{1}{|\mathcal{N}|}\sum_{\forall i\in\mathcal{N}}}_{\text{information fusion}}\underbrace{\left(\mathrm{f}(\hat{\mathbf{x}}_{n-1})+|\mathcal{N}|\mathbf{G}_{i,n}\Big(\mathbf{y}_{i,n}-\mathrm{h}_{i}\left(\mathrm{f}\left(\hat{\mathbf{x}}_{n-1}\right)\right)\right)}_{\text{local filtering operation}}.\nonumber
\end{align}
The expression in \eqref{eq:Midway-Term} essentially partitions the operation of \eqref{eq:Central-Filter} into two processes; i) local filtering operations using the observation information available at the individual agent, labelled as ``local filtering operation'' in \eqref{eq:Midway-Term}, ii) an information fusion operation that fusses local learning operations into a cohesive global operation, labelled as ``information fusion'' in \eqref{eq:Midway-Term}. 

\noindent\textbf{Remark~1:} The formulation in \eqref{eq:Midway-Term} forms the cornerstone of federated learning techniques~\cite{TAC,DistL,FedLearn}. In addition, note that only agents with new data sets at each time instant have to send updates to the fusion centre, while agents that have no new data sets only implement the projection operation. Moreover, subsets of agents can be selected to communicate with the fusion centre to avoid communication traffic. However, all these scenarios are special cases of the presented framework.

\subsection{From Federated to Distributed Learning}

In distributed learning scenarios, agents are connected through an ad-hoc network, that is modelled as a strongly connected graph. In this setting, agent $l$ can only communicate with its neighbouring agents denoted as $\mathcal{N}_{l}$, that includes agent $l$ itself. Thus, the information fusion step in \eqref{eq:Midway-Term} is approximated in a distributed fashion as
\begin{subequations}
\begin{align}
\boldsymbol{\phi}_{i,n}=&\mathrm{f}(\hat{\mathbf{x}}_{i,n-1})+\tilde{\mathbf{G}}_{i,n}\Big(\mathbf{y}_{i,n}-\mathrm{h}_{i}\left(\mathrm{f}\left(\hat{\mathbf{x}}_{i,n-1}\right)\right)\Big)\label{eq:Agent-Filter}
\\
\hat{\mathbf{x}}_{i,n}=&\boldsymbol{\phi}_{i,n}+\sum_{\forall j\in\mathcal{N}_{i}}c_{i,j}\left(\boldsymbol{\phi}_{i,n}-\boldsymbol{\phi}_{j,n}\right)\label{eq:Fuse}
\end{align} 
\label{eq:Local-Filtering-Operation}
\end{subequations}
where agent $i\in\mathcal{N}$ implements a local filtering operation, i.e., \eqref{eq:Agent-Filter}, using its local gain matrix $\tilde{\mathbf{G}}_{i,n}$, yielding the intermediate estimate $\boldsymbol{\phi}_{i,n}$; then, it aims to approximate the information fusion operation in \eqref{eq:Midway-Term} whereby intermediate estimates are fused with those of its neighbours in \eqref{eq:Fuse} using combination weights $\{c_{i,j}:\forall j\in\mathcal{N}_{i}\}$ which yields the final estimate, $\hat{\mathbf{x}}_{i,n}$. 

\vspace{0.08cm}

\noindent\textbf{Assumption~3:}
In keeping with convention, for both diffusion~\cite{diffusion} and consensus~\cite{average-consensus-journal,ConsFilter} frameworks, it is assumed that the network remains connected\footnote{That is, there exists an path between any two agents.} and 
\begin{equation}
\forall i\in\mathcal{N}:\hspace{0.12cm}\sum_{\forall j\in\mathcal{N}_{i}}c_{i,j}=1.
\label{eq:Convex-Combination}
\end{equation}
Therefore, matrix $\mathbf{C}$ with $i^{\text{th}}$ row and $j^{\text{th}}$ column element	
\[
\mathbf{C}^{\{i,j\}}=\left\{\begin{matrix}c_{i,j}&{}&\text{if}\hspace{0.12cm}j\in\mathcal{N}_{i}\\0&{}&\text{otherwise}\end{matrix}\right.
\]
will be row-stochastic and primitive~\cite{SayedBook}. 

\vspace{0.08cm}

\noindent\textbf{Remark~2:} There are two main techniques for fully distributed information fusion, diffusion~\cite{diffusion} and consensus~\cite{ConsFilter,average-consensus-journal} with the formulation in \eqref{eq:Midway-Term} accommodating both. 

\section{Performance Discrepancy and Stability}

The estimation error at agent $i$ is defined as 
\begin{equation}
\boldsymbol{\epsilon}_{i,n}=\mathbf{x}_{n}-\hat{\mathbf{x}}_{i,n}
\label{eq:Agent-Error-Def}
\end{equation}
while the network-wide perspective of the estimation error of all agents is given by
\begin{equation}
\mathbfcal{E}_{n}=\begin{bmatrix}\boldsymbol{\epsilon}^{\T}_{1,n},\ldots,\boldsymbol{\epsilon}^{\T}_{|\mathcal{N}|,n}\end{bmatrix}^{\T}.
\label{eq:Net-Error-Term}
\end{equation}
Taking performance of the centralised adaptation process in \eqref{eq:Central-Filter} as the benchmark, performance discrepancy between \eqref{eq:Central-Filter} and \eqref{eq:Local-Filtering-Operation} can now be formulated as
\begin{align}
\boldsymbol{\Delta}_{n+1}=&\mathbfcal{E}_{n+1}-\mathbf{1}\otimes\boldsymbol{\epsilon}_{n+1}\label{eq:Error-Dif}
\\
=&\frac{1}{|\mathcal{N}|}\textbf{\emph{1}}\begin{bmatrix}\mathrm{f}(\hat{\mathbf{x}}_{n})\\ \vdots \\ \mathrm{f}(\hat{\mathbf{x}}_{n})\end{bmatrix}-\mathbfcal{C}\begin{bmatrix}\mathrm{f}(\hat{\mathbf{x}}_{1,n})\\ \vdots \\ \mathrm{f}(\hat{\mathbf{x}}_{|\mathcal{N}|,n})\end{bmatrix}\nonumber
\\
&+\left(\frac{1}{|\mathcal{N}|}\textbf{\emph{1}}\mathbfcal{G}_{n+1}-\mathbfcal{C}\tilde{\mathbfcal{G}}_{n+1}\right)\mathbf{w}_{n+1}\nonumber
\\
&+\left(\frac{1}{|\mathcal{N}|}\textbf{\emph{1}}\mathbfcal{G}_{n+1}-\mathbfcal{C}\tilde{\mathbfcal{G}}_{n+1}\right)\mathrm{h}\left(\mathrm{f}(\mathbf{x}_{n},\mathbf{v}_{n})\right)\nonumber
\\	&-\frac{1}{|\mathcal{N}|}\textbf{\emph{1}}\mathbfcal{G}_{n+1}\begin{bmatrix}\mathrm{h}_{1}\left(\mathrm{f}(\hat{\mathbf{x}}_{n})\right)\\ \vdots \\ \mathrm{h}_{|\mathcal{N}|}\left(\mathrm{f}(\hat{\mathbf{x}}_{n})\right)\end{bmatrix}+\mathbfcal{C}\tilde{\mathbfcal{G}}_{n+1}\begin{bmatrix}\mathrm{h}_{1}\left(\mathrm{f}(\hat{\mathbf{x}}_{1,n})\right)\\ \vdots \\ \mathrm{h}_{|\mathcal{N}|}\left(\mathrm{f}(\hat{\mathbf{x}}_{|\mathcal{N}|,n})\right)\end{bmatrix}\nonumber
\end{align}
where $\textbf{\emph{1}}=\mathbf{1}\mathbf{1}^{\T}\otimes\mathbf{I}$, $\mathbfcal{G}_{n}=\diag{|\mathcal{N}|\mathbf{G}_{i,n}:\forall i\in\mathcal{N}}$, and $\tilde{\mathbfcal{G}}_{n}=\diag{\tilde{\mathbf{G}}_{i,n}:\forall i\in\mathcal{N}}$. Note that the derivation of \eqref{eq:Error-Dif} is given in Supplementary Material SM-1~\cite{SM}. The expression in \eqref{eq:Error-Dif} shows performance discrepancy between distributed and centralised approaches in relation to the network typology, gain matrices, and observation/evolution~functions. 

\subsection{The Case of Federated Learning}

In order to formulate and isolate the effect of replacing federated information fusion in \eqref{eq:Midway-Term} with that of the distributed information fusion step in \eqref{eq:Fuse}, it is assumed that both approaches use the equivalent gain matrices, that is,  
\[
\forall i\in\mathcal{N}:\hspace{0.12cm}\tilde{\mathbf{G}}_{i,n}=|\mathcal{N}|\mathbf{G}_{i,n}.
\]
Now, from \eqref{eq:Error-Dif} and $\tilde{\mathbfcal{G}}_{n+1}$ one has
\begin{align}
\boldsymbol{\Delta}_{n+1}=&\mathbfcal{C}\left(\begin{bmatrix}\mathrm{f}(\hat{\mathbf{x}}_{n})\\ \vdots \\ \mathrm{f}(\hat{\mathbf{x}}_{n})\end{bmatrix}-\begin{bmatrix}\mathrm{f}(\hat{\mathbf{x}}_{1,n})\\ \vdots \\ \mathrm{f}(\hat{\mathbf{x}}_{|\mathcal{N}|,n})\end{bmatrix}\right)\label{eq:Balanced-Gain-Matrix}
\\
&-\mathbfcal{C}\tilde{\mathbfcal{G}}_{n+1}\left(\begin{bmatrix}\mathrm{h}_{1}\left(\mathrm{f}(\hat{\mathbf{x}}_{n})\right)\\ \vdots \\ \mathrm{h}_{|\mathcal{N}|}\left(\mathrm{f}(\hat{\mathbf{x}}_{n})\right)\end{bmatrix}-\begin{bmatrix}\mathrm{h}_{1}\left(\mathrm{f}(\hat{\mathbf{x}}_{1,n})\right)\\ \vdots \\ \mathrm{h}_{|\mathcal{N}|}\left(\mathrm{f}(\hat{\mathbf{x}}_{|\mathcal{N}|,n})\right)\end{bmatrix}\right)\nonumber
\\
&+\left(\frac{1}{|\mathcal{N}|}\textbf{\emph{1}}-\mathbfcal{C}\right)\tilde{\mathbfcal{G}}_{n+1}\begin{bmatrix}\mathrm{h}_{1}\left(\mathrm{f}(\mathbf{x}_{n},\mathbf{v}_{n})\right)\\ \vdots \\ \mathrm{h}_{|\mathcal{N}|}\left(\mathrm{f}(\mathbf{x}_{n},\mathbf{v}_{n})\right)\end{bmatrix}\nonumber
\\
&-\left(\frac{1}{|\mathcal{N}|}\textbf{\emph{1}}-\mathbfcal{C}\right)\tilde{\mathbfcal{G}}_{n+1}\begin{bmatrix}\mathrm{h}_{1}\left(\mathrm{f}(\hat{\mathbf{x}}_{n})\right)\\ \vdots \\ \mathrm{h}_{|\mathcal{N}|}\left(\mathrm{f}(\hat{\mathbf{x}}_{n})\right)\end{bmatrix}\nonumber
\\
&+\left(\frac{1}{|\mathcal{N}|}\textbf{\emph{1}}-\mathbfcal{C}\right)\tilde{\mathbfcal{G}}_{n+1}\mathbf{w}_{n+1}\nonumber
\end{align}
with the full derivation given in Supplementary Material SM-2~\cite{SM}. Now, consider the case where at time instant $m\in\mathbb{N}$ there exists $\mu'>0$ so that $\|\hat{\mathbf{x}}_{m}-\hat{\mathbf{x}}_{i,m}\|^{2}\leq\mu'$
resulting in 
\begin{equation}
\forall i\in\mathcal{N}: \left\{\begin{aligned}&\mathrm{h}_{i}\left(\mathrm{f}\left(\hat{\mathbf{x}}_{m}\right)\right)-\mathrm{h}_{i}\left(\mathrm{f}\left(\hat{\mathbf{x}}_{i,m}\right)\right)=\mathbf{H}_{i}\mathbf{A}\boldsymbol{\Delta}_{m}+\mathrm{Res}_{\mathrm{h}_{i},\mathrm{f}}
\\
&
\\
&\mathrm{f}(\hat{\mathbf{x}}_{m})-\mathrm{f}(\hat{\mathbf{x}}_{i,m})=\mathbf{A}\boldsymbol{\Delta}_{m}+\mathrm{Res}_{\mathrm{f}}\end{aligned}\right.
\label{eq:Limit}
\end{equation}
where $\mathbf{A}$ and $\mathbf{H}_{i}$ denote the first-order approximation of $\mathrm{f}(\cdot)$ and $\mathrm{h}_{i}(\cdot)$, while $\mathrm{Res}_{\mathrm{f}}$ and $\mathrm{Res}_{\mathrm{h}_{i},\mathrm{f}}$ denote the residuals. From \eqref{eq:Balanced-Gain-Matrix}  and \eqref{eq:Limit} it follows that
\begin{align}
\E{\|\boldsymbol{\Delta}_{m+1}\|^{2}}\leq&\p{\mathbfcal{C}\left(\mathbf{I}-\tilde{\mathbfcal{G}}_{n+1}\mathbfcal{H}\right)\mathbfcal{A}}\E{\|\boldsymbol{\Delta}_{m}\|^{2}}\label{eq:MSDiff-Balanced}
\\
&+\p{\left(\frac{1}{|\mathcal{N}|}\textbf{\emph{1}}-\mathbfcal{C}\right)\tilde{\mathbfcal{G}}_{n+1}}\tr{{\boldsymbol{\Sigma}}_{\mathbf{w}}}\nonumber
\\
&+\|\mathrm{Res}_{\mathrm{f}}\|^{2}+\sum_{\forall i\in\mathcal{N}}\|\mathrm{Res}_{\mathrm{h}_{i},\mathrm{f}}\|^{2}\nonumber
\\
&+\p{\left(\frac{1}{|\mathcal{N}|}\textbf{\emph{1}}-\mathbfcal{C}\right)\tilde{\mathbfcal{G}}_{n+1}}\|\mathrm{Res}_{\text{ob}}\|^{2}\nonumber
\end{align}
where $\mathbfcal{H}=\diag{\mathbf{H}_{i}:\forall i\in\mathcal{N}}$, $\mathbfcal{A}=\mathbf{I}\otimes\mathbf{A}$, $\boldsymbol{\Sigma}_{\mathbf{w}}=\E{\mathbf{W}_{n}\mathbf{W}^{\T}_{n}}$,
\[
\mathrm{Res}_{\text{ob}}=\begin{bmatrix}\mathrm{h}_{1}\left(\mathrm{f}(\mathbf{x}_{n},\mathbf{v}_{n})\right)\\ \vdots \\ \mathrm{h}_{|\mathcal{N}|}\left(\mathrm{f}(\mathbf{x}_{n},\mathbf{v}_{n})\right)\end{bmatrix}-\begin{bmatrix}\mathrm{h}_{1}\left(\mathrm{f}(\hat{\mathbf{x}}_{n})\right)\\ \vdots \\ \mathrm{h}_{|\mathcal{N}|}\left(\mathrm{f}(\hat{\mathbf{x}}_{n})\right)\end{bmatrix}.
\]
This allows for the following results to be formulated

\vspace{0.08cm}

\noindent\textbf{Lemma~1:} If \textit{Assumption}~3  holds; then, 
\[
\p{\frac{1}{|\mathcal{N}|}\textbf{\emph{1}}-\mathbfcal{C}}<1.
\]

\noindent\textit{Proof of Lemma~1:} See the Supplementary Material, SM-3~\cite{SM}.

\vspace{0.08cm}

\noindent\textbf{Theorem~1:} If $\forall i\in\mathcal{N}$ the information set $\{\mathbf{y}_{j,m}:j\in\mathcal{N}_{i},m\leq n\}$ is sufficient to allow agent $i$ to find a gain matrix that satisfies \eqref{eq:Exponetially-Bounded}. Then, $\|\boldsymbol{\Delta}_{n}\|^{2}$ as given in \eqref{eq:MSDiff-Balanced} is bounded. 

\vspace{0.08cm}

\noindent\textit{Proof of Theorem~1:} See Supplementary Material SM-4~\cite{SM}.

\noindent\textbf{Remark~3:} The results in \textit{Theorem}~1, expand on previous works in this area~\cite{NonlinearDiffKalman,DEKF}. 

\noindent\textbf{Remark~3:} In general, \eqref{eq:MSDiff-Balanced} indicates that combination weights and gains should be selected in conjunction if convergence is to be achieved in wider scenarios. Although the results form \textit{Theorem}~1, expand on previous works in this aria~\cite{NonlinearDiffKalman,DEKF}, in the following section more general scenarios are considered.

\noindent\textbf{Remark~4:} From \eqref{eq:MSDiff-Balanced}, note that $\|\boldsymbol{\Delta}_{m}\|^{2}$ finds a minimum as $\mathbfcal{C}\rightarrow\frac{1}{|\mathcal{N}|}\textbf{\emph{1}}$. This can be achieved through the a consensus framework~\cite{ConsFilter,AVJ}, with a communication complexity penalty for any required level of accuracy~\cite{TAC,TACQ}. 

\subsection{The Case of Distributed Learning}

The aim of each agent, $i\in\mathcal{N}$, is to implement filtering operations in \eqref{eq:Agent-Filter}, where the matrix $\tilde{\mathbf{G}}_{i,n} $ is calculated so as to implement the fundamental filtering process so that 
\begin{equation}
\boldsymbol{\phi}_{i,n+1}=\E{\mathbf{x}_{n}|\hat{\mathbf{x}}_{j,m},\mathbf{y}_{i,m}:m\leq n-1,\forall j\in\mathcal{N}_{i}}.
\label{eq:Fundamrtal-Filter}
\end{equation} 
The formulation in \eqref{eq:Fundamrtal-Filter} accommodates estimates shared with the agent from its neighbourhood, that is, $\{\hat{\mathbf{x}}_{j,m}:m\leq n-1,\forall j\in\mathcal{N}_{i}\}$, that according to \eqref{eq:Local-Filtering-Operation} incorporate information from estimates of their neighbours, as well as, their observations. Thus, modelling how information is generated and flows through the network. 

\vspace{0.08cm}

\noindent\textbf{Theorem~2:}
If information set $\{\hat{\mathbf{x}}_{j,m},\mathbf{y}_{i,m}:m\leq n-1,\forall j\in\mathcal{N}_{i}\}$, yields gain matrix $\tilde{\mathbf{G}}_{i,n}$ so that 
\begin{equation}
\forall i\in\mathcal{N}:\hspace{0.32cm}\|\mathbf{x}_{n}-\boldsymbol{\phi}_{i,n}\|^{2}\leq\gamma_{i}\|\underbrace{\mathbf{x}_{n-1}-\hat{\mathbf{x}}_{i,n-1}}_{\boldsymbol{\epsilon}_{i,n-1}}\|^{2}
\label{eq:Beyound-Neibourhood}
\end{equation}
where $\gamma_{i}\in(0,1)$; then, $\|\boldsymbol{\Delta}_{n}\|^{2}$ becomes bounded.

\noindent\textit{Proof of Theorem~2:} See Supplementary Material SM-5~\cite{SM}.

\vspace{0.08cm}

\noindent\textbf{Remark~3:} There are similar results available for linear~\cite{AliDisKalman} and nonlinear~\cite{DEKF,NonlinearDiffKalman,AliDistLearningOne} distributed filtering techniques, where convergence criteria are either set at each agent having sufficient observation information to converge~\cite{DEKF,NonlinearDiffKalman} or this information to be shared with them through neighbouring agents~\cite{AliDisKalman}. However, \textit{Theorem}~2 is more comprehensive, as its setting allows introduction of more elaborate scenarios.

A more useful and widely encountered scenario is where the goal of each agent remains as given in \eqref{eq:Fundamrtal-Filter}, whereas the condition in \eqref{eq:Beyound-Neibourhood} only holds for a subset of elements in $\boldsymbol{\phi}_{i,n}$. This setting is formulated as
\begin{equation}
\boldsymbol{\phi}^{c}_{i,n}=\boldsymbol{\Upsilon}_{i}\boldsymbol{\phi}_{i,n}\hspace{0.24cm}\text{and}\hspace{0.24cm}\boldsymbol{\phi}^{c'}_{i,n}=\boldsymbol{\Upsilon}'_{i}\boldsymbol{\phi}_{i,n}
\label{eq:Def-Sub-Vect}
\end{equation}
where $\boldsymbol{\Upsilon}_{i}$ (\textit{cf}. $\boldsymbol{\Upsilon}'_{i}$) passes elements of $\boldsymbol{\phi}_{i,n}$ that meet (\textit{cf}. do not meet) the condition in \eqref{eq:Beyound-Neibourhood} unchanged, while zeroing other elements. From the definition in \eqref{eq:Def-Sub-Vect} it follows that 
\[
\forall i\in\mathcal{N}:\hspace{0.12cm} \boldsymbol{\phi}_{i,n}=\boldsymbol{\phi}^{c}_{i,n}+\boldsymbol{\phi}^{c'}_{i,n}\hspace{0.24cm}\text{and}\hspace{0.24cm}\boldsymbol{\Upsilon}_{i}+\boldsymbol{\Upsilon}'_{i}=\mathbf{I}
\]
and therefore, from \eqref{eq:Fuse}, we have
\begin{equation}
\boldsymbol{\Delta}_{n+1}=\mathbfcal{C}\begin{bmatrix}\hat{\mathbf{x}}_{n+1}-\boldsymbol{\phi}^{c}_{1,n+1}\\\vdots\\\hat{\mathbf{x}}_{n+1}-\boldsymbol{\phi}^{c}_{|\mathcal{N}|,n+1}\end{bmatrix}+\mathbfcal{C}\begin{bmatrix}\hat{\mathbf{x}}_{n+1}-\boldsymbol{\phi}^{c'}_{1,n+1}\\\vdots\\\hat{\mathbf{x}}_{n+1}-\boldsymbol{\phi}^{c'}_{|\mathcal{N}|,n+1}\end{bmatrix}.
\label{eq:Difference-First}
\end{equation}

\vspace{0.09cm}

\noindent\textbf{Assumption~4}: Since, the elements of $\boldsymbol{\phi}^{c'}_{i,n}$ are not observable at agent $i$, it is assumed that the local filtering operation in \eqref{eq:Agent-Filter} can be simplified to 
\[
\boldsymbol{\phi}^{c'}_{i,n}=\boldsymbol{\Upsilon}^{'}_{i}\mathrm{f}\left(\hat{\mathbf{x}}_{i,n-1}\right)
\]
that is, these elements are only assumed to be projections from the previous time instant, as there is not enough information in the observations to update them. 

\vspace{0.08cm}

\noindent\textbf{Theorem~3:} Given $g_{\boldsymbol{\Delta}_{m}}=\p{\mathbfcal{C}\left(\mathbf{I}-\boldsymbol{\Upsilon}\tilde{\mathbfcal{G}}_{m+1}\mathbfcal{H}\right)\mathbfcal{A}}$, if
\[
\exists k\in\mathbb{N}\hspace{0.12cm}\text{so~that}\hspace{0.12cm}\left(\prod^{k}_{i=0}g_{\boldsymbol{\Delta}_{m-i}}\right)<1
\]
then, $\|\boldsymbol{\Delta}_{n}\|^{2}$ becomes bounded.

\noindent\textit{Proof of Theorem~3:} See Supplementary Material SM-6~\cite{SM}.

\vspace{0.08cm}

\noindent\textbf{Remark~4:} \textit{Theorem~3} relates convergence to the information available to the agent, the dynamics of the system, that is, $\mathrm{f}(\cdot)$ and $\{\mathrm{h}_{i}(\cdot):i\in\mathcal{N}\}$, and the mechanism with which information flows through the network. In essence, this condition states that for each element of $\boldsymbol{\Phi}^{c'}_{i,n}$  that cannot be learnt from the information available to agent $i$, there must exist agents in the $k$-hop neighbourhood of agent $i$ to which this information is available. The parameter $k$ itself, determined by the state-space dynamics.

\section{Numerical Example}

For simulations, the network of $120$ agents were used with structure of the network as given in Supplementary Material SM-7~\cite{SM}.

\subsection{Statistical Learning}

\begin{figure}[h!]
\centering
\includegraphics[width=0.829\linewidth, trim = 0cm 3.6cm 0cm 0cm]{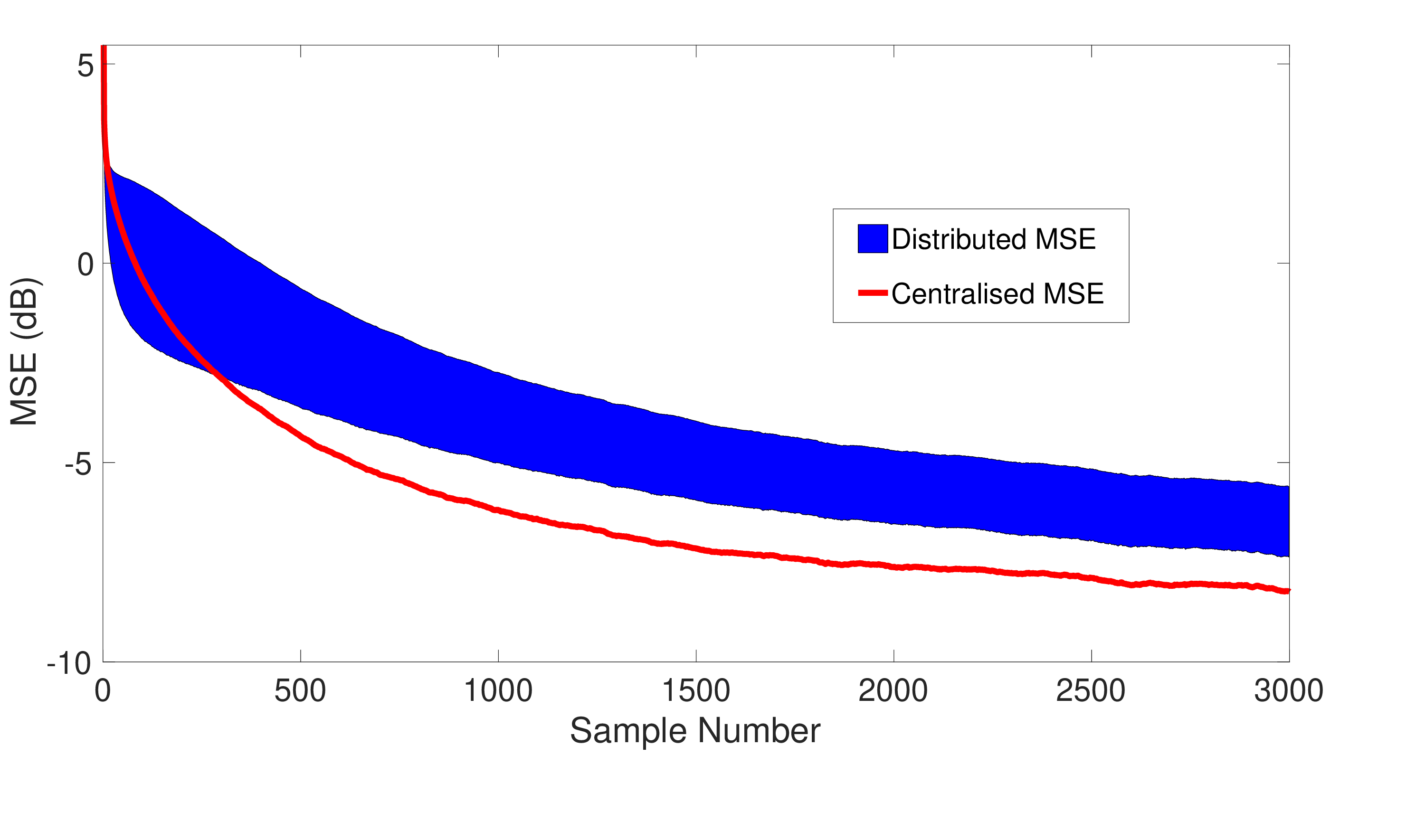}
\caption{MSE performance of distributed and centralised learning solutions. Mean averages were calculated over $2000$ independent realisations. The region in blue indicates where MSE of each agent lies for the distributed approach.}
\label{Fig:One-Converge}
\end{figure}

The most fundamental learning paradigm is considered, that is, each agent $i\in\mathcal{N}$ aims to learn the parameter $\hat{\mathbf{x}}_{i,n}$, so that 
\begin{equation}
\hat{\mathbf{y}}_{i,n}=\mathrm{h}_{i,n}(\hat{\mathbf{x}}_{i,n})=\varphi(\mathbf{Z}_{i,n}\hat{\mathbf{x}}_{i,n})
\label{eq:Stat-Learn-Example}
\end{equation}
is a best fit for observed information $\{\mathbf{y}_{i,m}:\forall i\in\mathcal{N},m<n\}$, where $\mathbf{Z}_{i,n}$ is a stochastic signal at agent $i$. This is achieved via solutions of the form in \eqref{eq:Local-Filtering-Operation}, where $\mathrm{f}(\cdot)$ is an identity function, $\mathrm{h}_{i,n}(\cdot)$ is formulated in \eqref{eq:Stat-Learn-Example}, and $
\tilde{\mathbf{G}}_{i,n}=\zeta\nabla_{\hat{\mathbf{x}}_{i,n}}\left(\|\mathbf{y}_{i,n}-\hat{\mathbf{y}}_{i,n}\|^{2}\right)$ with $\zeta\in(0,1)$. Agents seek the Pareto solution to the optimisation task 
\[
\min\E{\frac{1}{|\mathcal{N}|}\sum_{\forall i\in\mathcal{N}}\|\mathbf{y}_{i,n}-\hat{\mathbf{y}}_{i,n}\|^{2}}\cdot
\]
For simulations, $\{\mathbf{Z}_{i,n}:\forall i\in\mathcal{N},n\in\mathbb{N}\}$ was a set of zero-mean white Gaussian process and the nonlinearity, $\varphi(\cdot)$, used was the hyperbolic tangent function.  As a worst case scenario, it was considered that information required for convergence was only available to the collective of the agents That is, no one agent or neighbourhood of an gent had enough information for convergence. This falls beyond the convergence criteria set previously; however, the conditions satisfy \textit{Theorem}~3.  The MSE performance of each agent is benchmarked agents that of centralised learning in Fig.~\ref{Fig:One-Converge}. Notice that once more all agents exhibited convergent behaviour, validating the introduced concepts in this article.

\subsection{State-Space Filtering}

\begin{figure}[h!]
\centering
\includegraphics[width=0.82\linewidth,trim = 0cm 3.8cm 0cm 0cm]{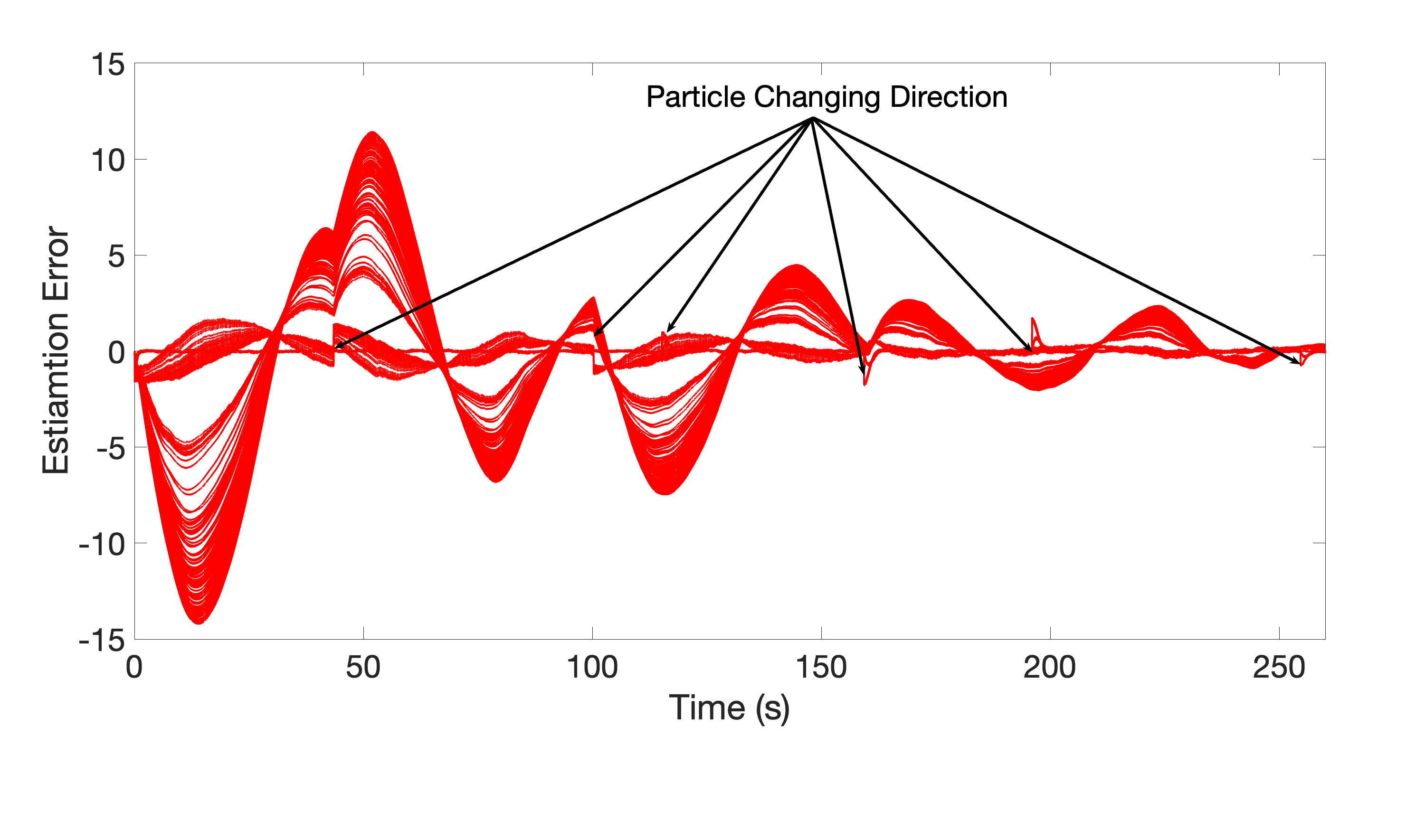}
\caption{Estimation error for the state of particle in the box. Estimation error of all elements of the state vector at all agents are shown.}
\label{Fig:PB}
\end{figure}

The problem of tracking a state-space system, with the nonlinear dynamics given in~\cite{targetnonlinear}, where the main objective was to track a target manoeuvring inside a two-dimensional box, was considered. The problem is referred to as ``\textit{particle-in-the-box}'' usually assuming measurements of the target location is available at each agent. However, in the simulation example here, it was assumed that only one agent had position measurements on the vertical axis, while other agents could only acquire position measurements on the horizontal axes. Note, that this setting presents a worst case scenario for distribute filtering. For the sake of simulations, the sampling rate was set at \unit[$25$]{Hz}, while the state evolution and observation noise covariances were $0.04\times\mathbf{I}$ and $0.16$. The distributed extended Kalman filter in~\cite{DEKF,NonlinearDiffKalman} was used to collaboratively track the target. Although these conditions fall outside the convergence requirements originally proposed in~\cite{DEKF,NonlinearDiffKalman}, the results in Fig.~\ref{Fig:PB} indicate all agents tracked the target with reasonable accuracy validating the theoretical concepts in \textit{Theorem}~3. 

\section{Conclusion}

The dynamics of distributed filtering and learning have been revisited from a novel perspective. The focus has been on the amount of information available to each agent, implementing iterative optimisation process in order to learn desired parameters from observed signals. This has allowed the formulation of a number of scenarios for distributed learning and filtering the convergence and stability of which falling beyond the scope of former studies. 

\newpage

\balance

\bibliographystyle{IEEEbib}

\bibliography{ref}


\newpage

\section*{Supplementary Material}

\subsection*{SM-1:Derivation of \eqref{eq:Error-Dif}}

From substituting \eqref{eq:State-Model} and \eqref{eq:Local-Filtering-Operation} into \eqref{eq:Agent-Error-Def}, one has
\[
\begin{aligned}
\boldsymbol{\epsilon}_{i,n+1}=&\mathrm{f}(\mathbf{x}_{n},\mathbf{v}_{n})-\sum_{\forall j\in\mathcal{N}_{i}}c_{i,j}\mathrm{f}(\hat{\mathbf{x}}_{j,n})
\\
&-\sum_{\forall j\in\mathcal{N}_{i}}c_{i,j}\tilde{\mathbf{G}}_{j,n+1}\Big(\mathbf{y}_{j,n+1}-\mathrm{h}_{j}\left(\mathrm{f}\left(\hat{\mathbf{x}}_{j,n}\right)\right)\Big).\nonumber
\end{aligned}
\]
that can be reformulated using \eqref{eq:Convex-Combination} to yield
\begin{align}
\boldsymbol{\epsilon}_{i,n+1}=&\sum_{\forall j\in\mathcal{N}_{i}}c_{i,j}\Big(\mathrm{f}(\mathbf{x}_{n},\mathbf{v}_{n})-\mathrm{f}(\hat{\mathbf{x}}_{j,n})\Big)\label{eq:Local-Error-Dynamic}
\\
&-\sum_{\forall j\in\mathcal{N}_{i}}c_{i,j}\tilde{\mathbf{G}}_{j,n+1}\Big(\mathrm{h}_{j}\left(\mathrm{f}\left(\mathbf{x}_{n},\mathbf{v}_{n}\right)\right)-\mathrm{h}_{j}\left(\mathrm{f}\left(\hat{\mathbf{x}}_{j,n}\right)\right)\Big)\nonumber
\\
&-\sum_{\forall j\in\mathcal{N}_{i}}c_{i,j}\tilde{\mathbf{G}}_{j,n+1}\mathbf{w}_{j,n+1}\nonumber.
\end{align}
Thus, the network-wide error in \eqref{eq:Net-Error-Term} can be expressed as
\begin{align}
\mathbfcal{E}_{n+1}=&\mathbfcal{C}\begin{bmatrix}\mathrm{f}(\mathbf{x}_{n},\mathbf{v}_{n})-\mathrm{f}(\hat{\mathbf{x}}_{1,n})\\ \vdots \\ \mathrm{f}(\mathbf{x}_{n},\mathbf{v}_{n})-\mathrm{f}(\hat{\mathbf{x}}_{|\mathcal{N}|,n})\end{bmatrix}-\mathbfcal{C}\tilde{\mathbfcal{G}}_{n+1}\mathbfcal{W}_{n+1}\label{eq:Net-Error-Final}
\\
&-\mathbfcal{C}\tilde{\mathbfcal{G}}_{n+1}\begin{bmatrix}\mathrm{h}_{1}\left(\mathrm{f}\left(\mathbf{x}_{n},\mathbf{v}_{n}\right)\right)-\mathrm{h}_{1}\left(\mathrm{f}\left(\hat{\mathbf{x}}_{1,n}\right)\right)\\ \vdots \\ \mathrm{h}_{|\mathcal{N}|}\left(\mathrm{f}\left(\mathbf{x}_{n},\mathbf{v}_{n}\right)\right)-\mathrm{h}_{|\mathcal{N}|}\left(\mathrm{f}\left(\hat{\mathbf{x}}_{|\mathcal{N}|,n}\right)\right)\end{bmatrix}\nonumber
\end{align}
where $\mathbfcal{C}=\mathbf{C}\otimes\mathbf{I}$, $\tilde{\mathbfcal{G}}_{n}=\diag{\tilde{\mathbf{G}}_{i,n}:\forall i\in\mathcal{N}}$, and 
\[
\mathbfcal{W}_{n}=\begin{bmatrix}\mathbf{w}^{\T}_{1,n},\ldots,\mathbf{w}^{\T}_{|\mathcal{N}|,n}\end{bmatrix}^{\T}.
\]
Now, deducting \eqref{eq:Cent-Error} from \eqref{eq:Net-Error-Final} gives
\begin{align}
\boldsymbol{\Delta}_{n+1}=&\mathbfcal{E}_{n+1}-\mathbf{1}\otimes\boldsymbol{\epsilon}_{n+1}\nonumber
\\
=&\frac{1}{|\mathcal{N}|}\textbf{\emph{1}}\begin{bmatrix}\mathrm{f}(\hat{\mathbf{x}}_{n})\\ \vdots\\ \mathrm{f}(\hat{\mathbf{x}}_{n})\end{bmatrix}-\mathbfcal{C}\begin{bmatrix}\mathrm{f}(\hat{\mathbf{x}}_{1,n})\\ \vdots \\ \mathrm{f}(\hat{\mathbf{x}}_{|\mathcal{N}|,n})\end{bmatrix}\nonumber
\\
&+\left(\frac{1}{|\mathcal{N}|}\textbf{\emph{1}}\mathbfcal{G}_{n+1}-\mathbfcal{C}\tilde{\mathbfcal{G}}_{n+1}\right)\mathbf{w}_{n+1}\nonumber
\\
&+\left(\frac{1}{|\mathcal{N}|}\textbf{\emph{1}}\mathbfcal{G}_{n+1}-\mathbfcal{C}\tilde{\mathbfcal{G}}_{n+1}\right)\mathrm{h}\left(\mathrm{f}(\mathbf{x}_{n},\mathbf{v}_{n})\right)\nonumber
\\
&-\frac{1}{|\mathcal{N}|}\textbf{\emph{1}}\mathbfcal{G}_{n+1}\begin{bmatrix}\mathrm{h}_{1}\left(\mathrm{f}(\hat{\mathbf{x}}_{n})\right)\\ \vdots \\ \mathrm{h}_{|\mathcal{N}|}\left(\mathrm{f}(\hat{\mathbf{x}}_{n})\right)\end{bmatrix}\nonumber
\\
&+\mathbfcal{C}\tilde{\mathbfcal{G}}_{n+1}\begin{bmatrix}\mathrm{h}_{1}\left(\mathrm{f}(\hat{\mathbf{x}}_{1,n})\right)\\ \vdots \\ \mathrm{h}_{|\mathcal{N}|}\left(\mathrm{f}(\hat{\mathbf{x}}_{|\mathcal{N}|,n})\right)\end{bmatrix}\nonumber
\end{align}
with $\mathbfcal{G}_{n}=\diag{|\mathcal{N}|\mathbf{G}_{i,n}:\forall i\in\mathcal{N}}$ and $\textbf{\emph{1}}=\mathbf{1}\mathbf{1}^{\T}\otimes\mathbf{I}$.

\subsection*{SM-2: Derivation of \eqref{eq:Balanced-Gain-Matrix}}

Substituting, all gain related matrices in \eqref{eq:Error-Dif} with $\tilde{\mathbfcal{G}}_{n+1}$ yields
\begin{align}
\boldsymbol{\Delta}_{n+1}=&\mathbfcal{C}\left(\begin{bmatrix}\mathrm{f}(\hat{\mathbf{x}}_{n})\\ \vdots \\ \mathrm{f}(\hat{\mathbf{x}}_{n})\end{bmatrix}-\begin{bmatrix}\mathrm{f}(\hat{\mathbf{x}}_{1,n})\\ \vdots \\ \mathrm{f}(\hat{\mathbf{x}}_{|\mathcal{N}|,n})\end{bmatrix}\right)\nonumber
\\
&-\frac{1}{|\mathcal{N}|}\textbf{\emph{1}}\tilde{\mathbfcal{G}}_{n+1}\begin{bmatrix}\mathrm{h}_{1}\left(\mathrm{f}(\hat{\mathbf{x}}_{n})\right)\\ \vdots \\ \mathrm{h}_{|\mathcal{N}|}\left(\mathrm{f}(\hat{\mathbf{x}}_{n})\right)\end{bmatrix}\nonumber
\\
&+\mathbfcal{C}\tilde{\mathbfcal{G}}_{n+1}\begin{bmatrix}\mathrm{h}_{1}\left(\mathrm{f}(\hat{\mathbf{x}}_{1,n})\right)\\ \vdots \\ \mathrm{h}_{|\mathcal{N}|}\left(\mathrm{f}(\hat{\mathbf{x}}_{|\mathcal{N}|,n})\right)\end{bmatrix}\nonumber
\\
&+\left(\frac{1}{|\mathcal{N}|}\textbf{\emph{1}}-\mathbfcal{C}\right)\tilde{\mathbfcal{G}}_{n+1}\mathbf{w}_{n+1}\nonumber
\\
&+\left(\frac{1}{|\mathcal{N}|}\textbf{\emph{1}}-\mathbfcal{C}\right)\tilde{\mathbfcal{G}}_{n+1}\mathrm{h}\left(\mathrm{f}(\mathbf{x}_{n},\mathbf{v}_{n})\right)\nonumber
\end{align}
which can be rearranged into
\begin{align}
\boldsymbol{\Delta}_{n+1}=&\mathbfcal{C}\left(\begin{bmatrix}\mathrm{f}(\hat{\mathbf{x}}_{n})\\ \vdots \\ \mathrm{f}(\hat{\mathbf{x}}_{n})\end{bmatrix}-\begin{bmatrix}\mathrm{f}(\hat{\mathbf{x}}_{1,n})\\ \vdots \\ \mathrm{f}(\hat{\mathbf{x}}_{|\mathcal{N}|,n})\end{bmatrix}\right)
\\
&-\mathbfcal{C}\tilde{\mathbfcal{G}}_{n+1}\left(\begin{bmatrix}\mathrm{h}_{1}\left(\mathrm{f}(\hat{\mathbf{x}}_{n})\right)\\ \vdots \\ \mathrm{h}_{|\mathcal{N}|}\left(\mathrm{f}(\hat{\mathbf{x}}_{n})\right)\end{bmatrix}-\begin{bmatrix}\mathrm{h}_{1}\left(\mathrm{f}(\hat{\mathbf{x}}_{1,n})\right)\\ \vdots \\ \mathrm{h}_{|\mathcal{N}|}\left(\mathrm{f}(\hat{\mathbf{x}}_{|\mathcal{N}|,n})\right)\end{bmatrix}\right)\nonumber
\\
&+\left(\frac{1}{|\mathcal{N}|}\textbf{\emph{1}}-\mathbfcal{C}\right)\tilde{\mathbfcal{G}}_{n+1}\begin{bmatrix}\mathrm{h}_{1}\left(\mathrm{f}(\mathbf{x}_{n},\mathbf{v}_{n})\right)\\ \vdots \\ \mathrm{h}_{|\mathcal{N}|}\left(\mathrm{f}(\mathbf{x}_{n},\mathbf{v}_{n})\right)\end{bmatrix}\nonumber
\\
&-\left(\frac{1}{|\mathcal{N}|}\textbf{\emph{1}}-\mathbfcal{C}\right)\tilde{\mathbfcal{G}}_{n+1}\begin{bmatrix}\mathrm{h}_{1}\left(\mathrm{f}(\hat{\mathbf{x}}_{n})\right)\\ \vdots \\ \mathrm{h}_{|\mathcal{N}|}\left(\mathrm{f}(\hat{\mathbf{x}}_{n})\right)\end{bmatrix}\nonumber
\\
&+\left(\frac{1}{|\mathcal{N}|}\textbf{\emph{1}}-\mathbfcal{C}\right)\tilde{\mathbfcal{G}}_{n+1}\mathbf{w}_{n+1}.\nonumber
\end{align}

\subsection*{SM-3: Proof of Lemma~1}

On one hand, note that
\[
\frac{1}{|\mathcal{N}|}\textbf{\emph{1}}-\mathbfcal{C}=\left(\frac{1}{|\mathcal{N}|}\mathbf{1}\mathbf{1}^{\T}-\mathbf{C}\right)\otimes\mathbf{I}
\]
and on the other hand, one has
\[
\left(\frac{1}{|\mathcal{N}|}\mathbf{1}\mathbf{1}^{\T}-\mathbf{C}\right)^{n}=(-1)^{n-1}\left(\frac{1}{|\mathcal{N}|}\mathbf{1}\mathbf{1}^{\T}-\mathbf{C}\right)\mathbf{C}^{n-1}.
\]
Moreover, from \textit{Assumption}~2, it follows that
\[
\forall \mathbf{a}: \exists a\hspace{0.2cm}\text{so that}\hspace{0.2cm}\lim_{n\rightarrow\infty}\mathbf{C}^{n}\mathbf{a}=a\mathbf{1}
\]
Therefore, 
\[
\begin{aligned}
\forall\mathbf{a}:\hspace{0.2cm}\lim_{n\rightarrow\infty}&\left(\frac{1}{|\mathcal{N}|}\mathbf{1}\mathbf{1}^{\T}-\mathbf{C}\right)^{n}\mathbf{a}
\\
=&\lim_{n\rightarrow\infty}(-1)^{n-1}\left(\frac{1}{|\mathcal{N}|}\mathbf{1}\mathbf{1}^{\T}-\mathbf{C}\right)\mathbf{C}^{n-1}\mathbf{a}
\\
=&\lim_{n\rightarrow\infty}(-1)^{n-1}\left(\frac{1}{|\mathcal{N}|}\mathbf{1}\mathbf{1}^{\T}-\mathbf{C}\right)(a\mathbf{1})=0
\end{aligned}
\]
indicating $\p{\frac{1}{|\mathcal{N}|}\mathbf{1}\mathbf{1}^{\T}-\mathbf{C}}<1$, and hence, $\p{\frac{1}{|\mathcal{N}|}\textbf{\emph{1}}-\mathbfcal{C}}<1$. Thus, concluding the proof.

\subsection*{SM-4: Proof of Theorem~1:}

From \textit{Lemma}~1 and given \eqref{eq:Exponetially-Bounded}, it follows that $\|\mathrm{Res}_{\text{ob}}\|^{2}$ will be bounded. Moreover,  conventionally, it is assumed that $\|\mathrm{Res}_{\mathrm{f}}\|^{2}$, $\|\mathrm{Res}_{\mathrm{h}_{i},\mathrm{f}}\|^{2}$, and $\tr{\boldsymbol{\Sigma}_{\mathbf{w}}}$ are also bounded~\cite{EKFS,NonlinearDiffKalman,DEKF}. Moreover, from the conditions of this theorem and by definition, 
\[
\forall i\in\mathcal{N}:\p{\sum_{\forall j\in\mathcal{N}_{i}}\left(\mathbf{I}-\tilde{\mathbf{G}}_{j,n}\mathbf{H}_{j}\right)\mathbf{A}}<1.
\]
Thus, given \textit{Assumption}~2, one has 
\[
\p{\mathbfcal{C}(\mathbf{I}-\tilde{\mathbfcal{G}}_{n}\mathbfcal{H})\mathbfcal{A}}<1
\]
concluding the proof.

\subsection*{SM-5: Proof of Theorem~2:}

Substituting \eqref{eq:Local-Filtering-Operation} into \eqref{eq:Error-Dif} gives
\[
\begin{aligned}
\boldsymbol{\Delta}_{n+1}=&\mathbfcal{E}_{n+1}-\mathbf{1}\otimes\boldsymbol{\epsilon}_{n+1}
\\
=&\begin{bmatrix}
\mathbf{x}_{n+1}-\sum_{\forall j\in\mathcal{N}_{1}}\boldsymbol{\phi}_{j,n+1}\\ \vdots\\ \mathbf{x}_{n+1}-\sum_{\forall j\in\mathcal{N}_{|N|}}\boldsymbol{\phi}_{j,n+1}\end{bmatrix}-\begin{bmatrix}\mathbf{x}_{n+1}-\hat{\mathbf{x}}_{n+1}\\ \vdots\\ \mathbf{x}_{n+1}-\hat{\mathbf{x}}_{n+1}\end{bmatrix}
\\
=&\mathbfcal{C}\begin{bmatrix}\hat{\mathbf{x}}_{n+1}-\boldsymbol{\phi}_{1,n+1}\\\vdots \\ \hat{\mathbf{x}}_{n+1}-\boldsymbol{\phi}_{|\mathcal{N}|,n+1}\end{bmatrix}
\end{aligned}
\]
which yields
\begin{equation}
\|\boldsymbol{\Delta}_{n+1}\|^{2}=\p{\mathbfcal{C}}\sum_{\forall i\in\mathcal{N}}\|\hat{\mathbf{x}}_{n+1}-\boldsymbol{\phi}_{i,n+1}\|^{2}.
\label{eq:Mid-Express}
\end{equation}
Substituting \eqref{eq:Beyound-Neibourhood} into \eqref{eq:Mid-Express} gives
\[
\|\boldsymbol{\Delta}_{n+1}\|^{2}\leq\p{\mathbfcal{C}}\gamma\sum_{\forall i\in\mathcal{N}}\|\hat{\mathbf{x}}_{n+1}-\boldsymbol{\phi}_{i,n+1}\|^{2}
\]
with $\gamma=\max\{\gamma_{i},\forall i\in\mathcal{N}\}$, and thus, it would transpire that 
\begin{equation}
\|\boldsymbol{\Delta}_{n+1}\|^{2}\leq\p{\mathbfcal{C}}\gamma\|\boldsymbol{\Delta}_{n}\|^{2}.
\label{eq:Mono-Gain-Ob}
\end{equation}
Given that $\p{\mathbfcal{C}}=1$ and $\gamma\in(0,1)$, \eqref{eq:Mono-Gain-Ob} indicates all agents will converge and $\|\boldsymbol{\Delta}_{n}\|^{2}$ becomes bounded. Hence, concluding the proof. 

\subsection*{SM-6: Proof of Theorem~3:}

From \eqref{eq:Difference-First} and \textit{Assumption}~4 one has
\begin{align}
\boldsymbol{\Delta}_{n+1}=&\mathbfcal{C}\begin{bmatrix}\mathrm{f}(\hat{\mathbf{x}}_{n})-\mathrm{f}(\hat{\mathbf{x}}_{1,n})\\\vdots\\\mathrm{f}(\hat{\mathbf{x}}_{n})-\mathrm{f}(\hat{\mathbf{x}}_{|\mathcal{N}|,n})\end{bmatrix}\label{eq:Total-Mid}
\\
&+\frac{1}{|\mathcal{N}|}\textbf{\emph{1}}\mathbfcal{G}_{n+1}\begin{bmatrix}\mathrm{h}_{1}\left(\mathrm{f}\left(\mathbf{x}_{n},\mathbf{v}_{n}\right)\right)-\mathrm{h}_{1}\left(\mathrm{f}\left(\hat{\mathbf{x}}_{n}\right)\right)\\ \vdots \\ \mathrm{h}_{|\mathcal{N}|}\left(\mathrm{f}\left(\mathbf{x}_{n},\mathbf{v}_{n}\right)\right)-\mathrm{h}_{|\mathcal{N}|}\left(\mathrm{f}\left(\hat{\mathbf{x}}_{n}\right)\right)\end{bmatrix}\nonumber
\\
&-\mathbfcal{C}\boldsymbol{\Upsilon}\tilde{\mathbfcal{G}}_{n+1}\begin{bmatrix}\mathrm{h}_{1}\left(\mathrm{f}\left(\mathbf{x}_{n},\mathbf{v}_{n}\right)\right)-\mathrm{h}_{1}\left(\mathrm{f}\left(\hat{\mathbf{x}}_{1,n}\right)\right)\\ \vdots \\ \mathrm{h}_{|\mathcal{N}|}\left(\mathrm{f}\left(\mathbf{x}_{n},\mathbf{v}_{n}\right)\right)-\mathrm{h}_{|\mathcal{N}|}\left(\mathrm{f}\left(\hat{\mathbf{x}}_{|\mathcal{N}|,n}\right)\right)\end{bmatrix}\nonumber
\end{align}
where $\boldsymbol{\Upsilon}=\diag{\boldsymbol{\Upsilon}_{i}:i=1,\ldots,|\mathcal{N}|}$.  The expression in \eqref{eq:Total-Mid} can be rearranged to give
\begin{align}
\boldsymbol{\Delta}_{n+1}=&\mathbfcal{C}\begin{bmatrix}\mathrm{f}(\hat{\mathbf{x}}_{n})-\mathrm{f}(\hat{\mathbf{x}}_{1,n})\\\vdots\\\mathrm{f}(\hat{\mathbf{x}}_{n})-\mathrm{f}(\hat{\mathbf{x}}_{|\mathcal{N}|,n})\end{bmatrix}+\mathrm{Res}_{\boldsymbol{\Delta}_{m}}\label{eq:Total-Mid-2}
\\
&-\mathbfcal{C}\boldsymbol{\Upsilon}\tilde{\mathbfcal{G}}_{n+1}\begin{bmatrix}\mathrm{h}_{1}\left(\mathrm{f}\left(\hat{\mathbf{x}}_{n}\right)\right)-\mathrm{h}_{1}\left(\mathrm{f}\left(\hat{\mathbf{x}}_{1,n}\right)\right)\\ \vdots \\ \mathrm{h}_{|\mathcal{N}|}\left(\mathrm{f}\left(\hat{\mathbf{x}}_{n}\right)\right)-\mathrm{h}_{|\mathcal{N}|}\left(\mathrm{f}\left(\hat{\mathbf{x}}_{|\mathcal{N}|,n}\right)\right)\end{bmatrix}\nonumber
\end{align}  
with $\mathbf{S}=\left(\frac{1}{|\mathcal{N}|}\textbf{\emph{1}}\mathbfcal{G}_{n+1}-\mathbfcal{C}\boldsymbol{\Upsilon}\tilde{\mathbfcal{G}}_{n+1}\right)$ so that 
\begin{equation}
\mathrm{Res}_{\boldsymbol{\Delta}_{n}}=\mathbf{S}\begin{bmatrix}\mathrm{h}_{1}\left(\mathrm{f}\left(\mathbf{x}_{n},\mathbf{v}_{n}\right)\right)-\mathrm{h}_{1}\left(\mathrm{f}\left(\hat{\mathbf{x}}_{n}\right)\right)\\ \vdots \\ \mathrm{h}_{|\mathcal{N}|}\left(\mathrm{f}\left(\mathbf{x}_{n},\mathbf{v}_{n}\right)\right)-\mathrm{h}_{|\mathcal{N}|}\left(\mathrm{f}\left(\hat{\mathbf{x}}_{n}\right)\right)\end{bmatrix}.
\label{eq:Res-Delta}
\end{equation}
Once more, consider $\exists m\in\mathbb{N}$ so that $\|\hat{\mathbf{x}}_{m}-\hat{\mathbf{x}}_{i,m}\|^{2}\leq\mu'$ resulting in the approximations in \eqref{eq:Limit} to hold, and hence, yielding
\begin{equation}
\E{\|\boldsymbol{\Delta}_{m+1}\|^{2}}\leq g_{\boldsymbol{\Delta}_{m}}\E{\|\boldsymbol{\Delta}_{m}\|^{2}}+\|\mathrm{Res}_{\boldsymbol{\Delta}_{m}}\|^{2}.
\label{eq:MSDiff-NoHolds}
\end{equation}
Through $k$ iteration, \eqref{eq:MSDiff-NoHolds} gives
\begin{equation}
\begin{aligned}
\E{\|\boldsymbol{\Delta}_{m+1}\|^{2}}\leq&\left(\prod^{k}_{i=0}g_{\boldsymbol{\Delta}_{m-i}}\right)\E{\|\boldsymbol{\Delta}_{m-k}\|^{2}}
\\
&+\sum^{k}_{i=0}\left(\prod^{i-1}_{j=0}g_{\boldsymbol{\Delta}_{m-j}}\right)\|\mathrm{Res}_{\boldsymbol{\Delta}_{m-k}}\|^{2}
\\
&+\|\mathrm{Res}_{\boldsymbol{\Delta}_{m}}\|^{2}.
\end{aligned}
\label{eq:MSDiff-NoHold-I}
\end{equation}
From \eqref{eq:Res-Delta}, it follows that $\|\mathrm{Res}_{\boldsymbol{\Delta}_{m}}\|^{2}$ is bounded. Thus, from \textit{Assumption}~2~and~4 it follows that the expression in \eqref{eq:MSDiff-NoHold-I} converges, if 
\[
\exists k\in\mathbb{N}: \left(\prod^{k}_{i=0}g_{\boldsymbol{\Delta}_{m-i}}\right)<1
\]
which concludes the proof.

\subsection*{SM-7: Network used in Simulation Examples}

\begin{figure}[H]
\centering
\includegraphics[width=0.9\linewidth]{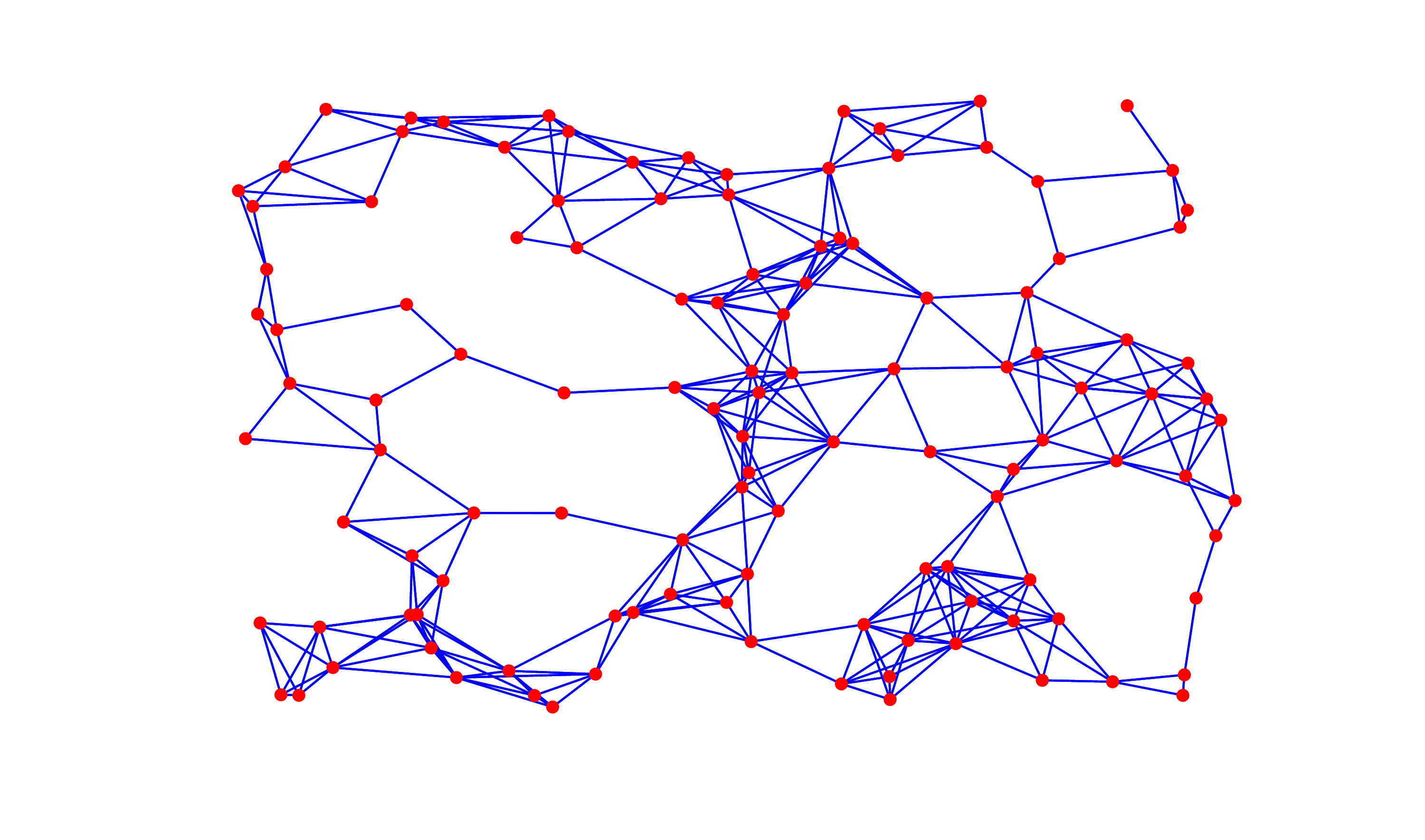}
\caption{Network of $120$ nodes and $352$ connections used in simulation examples.} 
\label{Fig:Network}
\end{figure}

Fig.~\ref{Fig:Network} shows typology of the network used in all simulation examples. In keeping with the literature conversion~\cite{SayedBook,AliDistLearningOne,DistL} all connections are assumed bidirectional.

\end{document}